\documentclass[11pt,a4paper]{article}
\usepackage[hyperref]{emnlp-ijcnlp-2019}
\usepackage{times}
\usepackage{latexsym}

\usepackage{url}

\usepackage{epsfig}
\usepackage{graphicx}
\usepackage{amsmath}
\usepackage{amssymb}

\usepackage{amsmath}
\usepackage{amsfonts}
\usepackage{multirow}
\usepackage{booktabs}

\aclfinalcopy 

\newcommand\NLVR{$\text{NLVR}^2$}
\newcommand\NL{ $\textbf{N}_\textbf{L}$}
\newcommand\NX{ $\textbf{N}_\textbf{X}$}
\newcommand\NR{ $\textbf{N}_\textbf{R}$ }

\title{LXMERT: Learning Cross-Modality Encoder Representations \\ from Transformers}

\author{Hao Tan \;\;\;\;\;\;\; Mohit Bansal \\
  UNC Chapel Hill \\
  {\tt \{haotan, mbansal\}@cs.unc.edu} \\
 }

\date{}

\begin{document}
\maketitle
\begin{abstract}
Vision-and-language reasoning requires an understanding of visual concepts, language semantics, and, most importantly, the alignment and relationships between these two modalities.
We thus propose the LXMERT (Learning Cross-Modality Encoder Representations from Transformers) framework to learn these vision-and-language connections. In LXMERT, we build a large-scale Transformer model that consists of three encoders: an object relationship encoder, a language encoder, and a cross-modality encoder.
Next, to endow our model with the capability of connecting vision and language semantics, we pre-train the model with large amounts of image-and-sentence pairs, via five diverse representative pre-training tasks: masked language modeling, masked object prediction (feature regression and label classification), cross-modality matching, and image question answering. 
These tasks help in learning both intra-modality and cross-modality relationships.
After fine-tuning from our pre-trained parameters, our model achieves the state-of-the-art results on two visual question answering datasets (i.e., VQA and GQA).
We also show the generalizability of our pre-trained cross-modality model by adapting it to a challenging visual-reasoning task, $\text{NLVR}^2$, and improve the previous best result by $22\%$ absolute ($54\%$ to $76\%$).
Lastly, we demonstrate detailed ablation studies to prove that both our novel model components and pre-training strategies significantly contribute to our strong results; and also present several attention visualizations for the different encoders.\footnote{Published at EMNLP 2019. Code and pre-trained models publicly available at:  \href{https://github.com/airsplay/lxmert}{https://github.com/airsplay/lxmert}}

\end{abstract}

\section{Introduction}
Vision-and-language reasoning requires the understanding of visual contents, language semantics, and cross-modal alignments and relationships.
There has been substantial past works in separately developing backbone models with better representations for the single modalities of vision and of language.
For visual-content understanding, people have developed several backbone models~\cite{simonyan2014very, szegedy2015going, he2016deep}
and shown their effectiveness on large vision datasets~\cite{deng2009imagenet, lin2014microsoft, krishna2017visual}. 
Pioneering works~\cite{girshick2014rich, xu2015show} also show the generalizability of these pre-trained (especially on ImageNet) backbone models by fine-tuning them on different tasks.
In terms of language understanding, last year, we witnessed strong progress towards building a universal backbone model with large-scale contextualized language model pre-training~\cite{peters2018deep, radford2018improving, devlin2018bert}, which has improved performances on various tasks~\cite{rajpurkar2016squad, wang2018glue} to significant levels.
Despite these influential single-modality works, large-scale pretraining and fine-tuning studies for the modality-pair of vision and language are still under-developed. 

Therefore, we present one of the first works in building a pre-trained vision-and-language cross-modality framework and show its strong performance on several datasets. 
We name this framework ``LXMERT: Learning Cross-Modality Encoder Representations from Transformers'' (pronounced: `leksmert'). 
This framework is modeled after recent BERT-style innovations while further adapted to useful cross-modality scenarios.
Our new cross-modality model focuses on learning vision-and-language interactions, especially for representations of a single image and its descriptive sentence.
It consists of three Transformer~\cite{vaswani2017attention} encoders: 
an object relationship encoder, a language encoder, and a cross-modality encoder.
In order to better learn the cross-modal alignments between vision and language, we next pre-train our model with five diverse representative tasks: (1) masked cross-modality language modeling, (2) masked object prediction via RoI-feature regression, (3) masked object prediction via detected-label classification, (4) cross-modality matching, and (5) image question answering.
Different from single-modality pre-training (e.g., masked LM in BERT), this multi-modality pre-training allows our model to infer masked features either from the visible elements in the same modality, or from aligned components in the other modality. 
In this way, it helps build both intra-modality and cross-modality relationships.

Empirically, we first evaluate LXMERT on two popular visual question-answering datasets, VQA~\cite{antol2015vqa} and GQA~\cite{hudson2019gqa}. 
Our model outperforms previous works in all question categories (e.g., Binary, Number, Open) and achieves state-of-the-art results in terms of overall accuracy. 
Further, to show the generalizability of our pre-trained model, we fine-tune LXMERT on a challenging visual reasoning task, Natural Language for Visual Reasoning for Real ($\text{NLVR}^2$)~\cite{suhr2018corpus}, where we do not use the natural images in their dataset for our pre-training, but fine-tune and evaluate on these challenging, real-world images.
In this setup, we achieve a large improvement of $22\%$ absolute in accuracy ($54\%$ to $76\%$, i.e., 48\% relative error reduction) and $30\%$ absolute in consistency ($12\%$ to $42\%$, i.e., 34\% relative error reduction).
Lastly, we conduct several analysis and ablation studies to prove the effectiveness of our model components and diverse pre-training tasks by removing them or comparing them with their alternative options. Especially, we use several ways to take the existing BERT model and its variants, and show their ineffectiveness in vision-and-language tasks, which overall proves the need of our new cross-modality pre-training framework. We  also present several attention visualizations for the different language, object-relationship, and cross-modality encoders.

\begin{figure*}[t]
\centering
\includegraphics[width=0.98\textwidth]{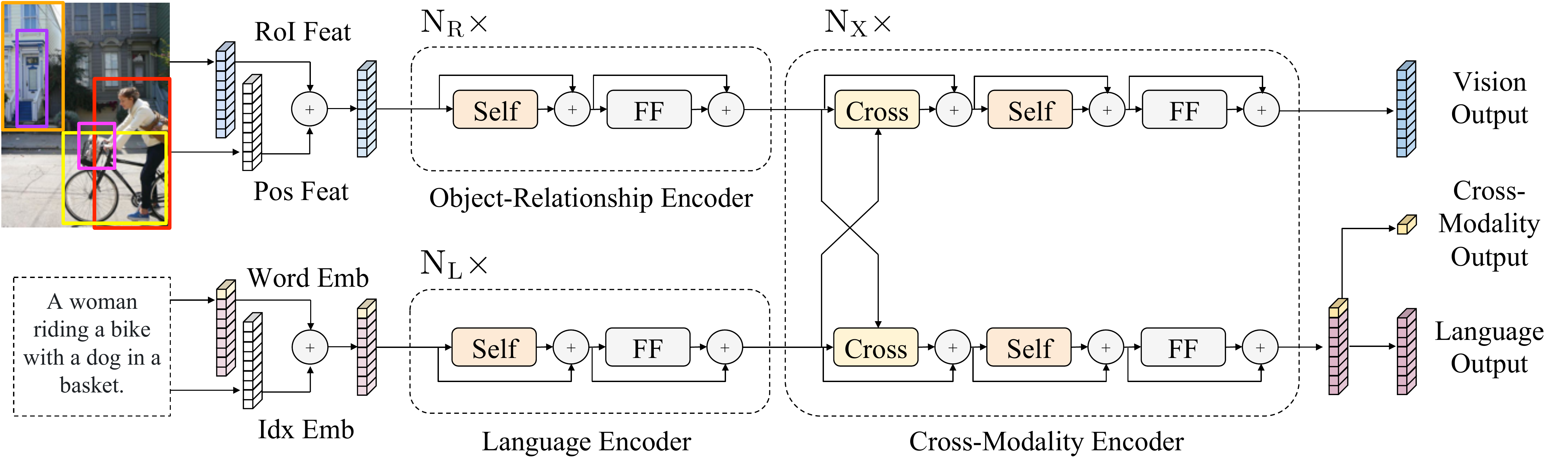}
\caption{
The LXMERT model for learning vision-and-language cross-modality representations. `Self' and `Cross' are abbreviations for self-attention sub-layers and cross-attention sub-layers, respectively. `FF' denotes a feed-forward sub-layer. 
}
\vspace{-5pt}
\label{fig:model}
\end{figure*}

\section{Model Architecture}
\label{sec:model}
We build our cross-modality model with self-attention and cross-attention layers following the recent progress in designing natural language processing models (e.g., transformers~\cite{vaswani2017attention}).
As shown in Fig.~\ref{fig:model}, our model takes two inputs: an image and its related sentence (e.g., a caption or a question). 
Each image is represented as a sequence of objects, and each sentence is represented as a sequence of words.
Via careful design and combination of these self-attention and cross-attention layers, our model is able to generate language representations, image representations, and cross-modality representations from the inputs.
Next, we describe the components of this model in detail.
\subsection{Input Embeddings}
\label{sec:input}
The input embedding layers in LXMERT convert the inputs (i.e., an image and a sentence) into two sequences of features: word-level sentence embeddings and object-level image embeddings.
These embedding features will be further processed by the latter encoding layers.
\paragraph{Word-Level Sentence Embeddings}
\label{sec:word_emb}
A sentence is first split into words $\left\{w_1, \ldots, w_n\right\}$ with length of $n$ by the same WordPiece tokenizer~\cite{wu2016google} in~\newcite{devlin2018bert}.
Next, as shown in Fig.~\ref{fig:model}, the word $w_i$ and its index $i$ ($w_i$'s absolute position in the sentence) are projected to vectors by embedding sub-layers, and then added to the index-aware word embeddings:
\begin{align*}
    \hat{w}_i &= \mathrm{WordEmbed}\left(w_i\right) \\
    \hat{u}_i & = \mathrm{IdxEmbed}\left(i\right) \\
    h_i & = \mathrm{LayerNorm}\left(\hat{w}_i + \hat{u}_i\right) 
\end{align*}

\paragraph{Object-Level Image Embeddings}
\label{sec:obj_emb}
Instead of using the feature map output by a convolutional neural network, we follow \newcite{anderson2018bottom} in taking the features of detected objects as the embeddings of images.
Specifically, the object detector detects $m$ objects $\left\{o_1, \ldots, o_m\right\}$ from the image (denoted by bounding boxes on the image in Fig.~\ref{fig:model}).
Each object $o_j$ is represented by its position feature (i.e., bounding box coordinates) $p_j$ and its $2048$-dimensional region-of-interest (RoI) feature $f_j$.
Instead of directly using the RoI feature $f_j$ without considering its position $p_j$
in \newcite{anderson2018bottom}, we learn a position-aware embedding $v_j$ by adding outputs of 2 fully-connected layers:
\begin{align}
    \hat{f}_j & = \mathrm{LayerNorm}\left(W_\textsc{f} f_j + b_\textsc{f} \right) \nonumber \\ 
    \hat{p}_j & = \mathrm{LayerNorm}\left(W_\textsc{p} p_j + b_\textsc{p} \right) \nonumber \\
    v_j & = \left(\hat{f}_j + \hat{p}_j\right) / 2 
    \label{eqn:object_emb}
\end{align}
In addition to providing spatial information in visual reasoning, 
the inclusion of positional information is necessary for our masked object prediction pre-training task (described in Sec.~\ref{sec:vision_task}).
Since the image embedding layer and the following attention layers are agnostic to the absolute indices of their inputs, the order of the object is not specified.
Lastly, in Equation \ref{eqn:object_emb}, the layer normalization is applied to the projected features before summation so as to balance the energy of the two different types of features.

\subsection{Encoders}
\label{sec:encoder}
We build our encoders, i.e., the language encoder, the object-relationship encoder, and the cross-modality encoder, mostly on the basis of two kinds of attention layers: self-attention layers and cross-attention layers. We first review the definition and notations of attention layers and then discuss how they form our encoders.
\paragraph{Background: Attention Layers} 
\label{sec:att}
Attention layers~\cite{bahdanau2014neural, xu2015show} aim to retrieve information from a set of \emph{context} vectors $\{y_j\}$ related to a \emph{query} vector $x$.
An attention layer first calculates the matching score $a_j$ between the \emph{query} vector $x$ and each \emph{context} vector $y_j$. Scores are then normalized by softmax:
\begin{align*}
\vspace{-5pt}
    a_j &= \mathrm{score}(x, y_j)    \\
    \alpha_j &= \exp(a_j) / \sum\nolimits_k \exp(a_k) 
\end{align*}
The output of an attention layer is the weighted sum of the \emph{context} vectors w.r.t. the softmax-normalized score:
$\mathrm{Att}_{\textsc{x}\rightarrow\textsc{y}}\left(x, \{y_j\}\right) = \sum\nolimits_j \alpha_j y_j$.
An attention layer is called \emph{self-attention} when the \emph{query} vector $x$ is in the set of \emph{context} vectors $\{y_j\}$.
Specifically, we use the multi-head attention following Transformer~\cite{vaswani2017attention}.

\paragraph{Single-Modality Encoders}
\label{sec:selfatt}
After the embedding layers, we first apply two transformer encoders~\cite{vaswani2017attention}, i.e., a \textbf{language encoder} and an \textbf{object-relationship encoder}, and each of them only focuses on a single modality (i.e., language or vision).
Different from BERT~\cite{devlin2018bert}, which applies the transformer encoder only to language inputs, we apply it to vision inputs as well (and to cross-modality inputs as described later below). 
Each layer (left dashed blocks in Fig.~\ref{fig:model}) in a single-modality encoder contains a self-attention (`Self') sub-layer and a feed-forward (`FF') sub-layer, where the feed-forward sub-layer is further composed of two fully-connected sub-layers.
We take $\textbf{N}_\textbf{L}$ and $\textbf{N}_\textbf{R}$ layers in the language encoder and the object-relationship encoder, respectively. 
We add a residual connection and layer normalization (annotated by the `+' sign in Fig.~\ref{fig:model}) after each sub-layer as in \newcite{vaswani2017attention}.

\begin{figure*}[t]
\centering
\includegraphics[width=0.98\textwidth]{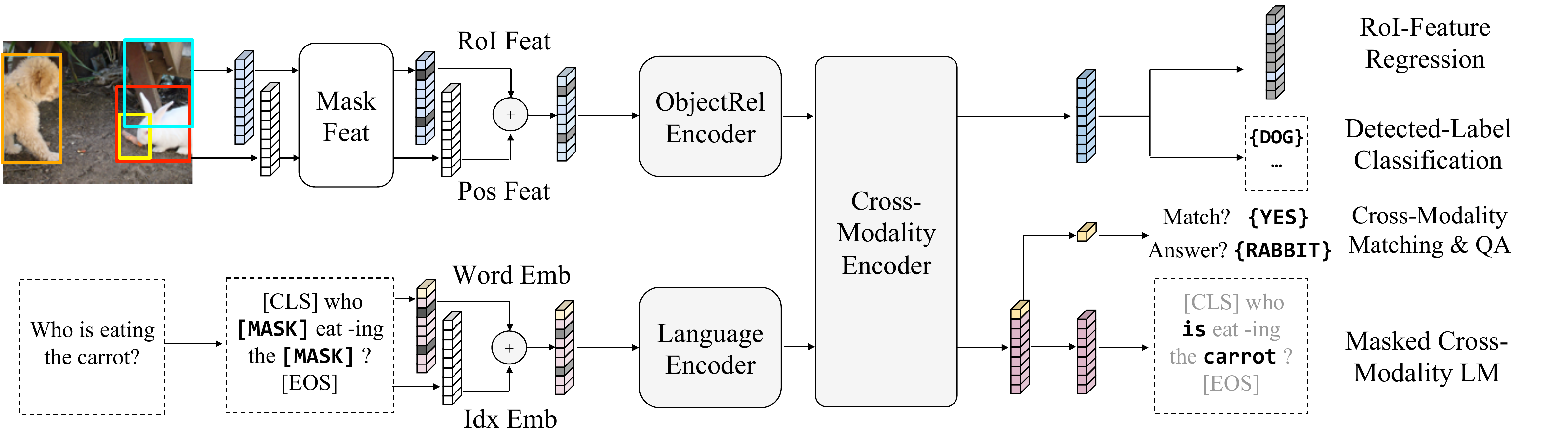}
\caption{
Pre-training in LXMERT. The object RoI features and word tokens are masked.
Our five pre-training tasks learn the feature representations based on these masked inputs.
Special tokens are in brackets and classification labels are in braces. 
}
\label{fig:pretrain}
\vspace{-8pt}
\end{figure*}

\paragraph{Cross-Modality Encoder}
\label{sec:crossatt}
Each cross-modality layer (the right dashed block in Fig.~\ref{fig:model}) in the cross-modality encoder consists of two self-attention sub-layers, one bi-directional cross-attention sub-layer, and two feed-forward sub-layers. 
We stack (i.e., using the output of $k$-th layer as the input of $(k\mbox{+}1)$-th layer) $\textbf{N}_\textbf{X}$ these cross-modality layers in our encoder implementation.
Inside the $k$-th layer, the bi-directional cross-attention sub-layer (`Cross') is first applied, which contains two uni-directional cross-attention sub-layers: one from language to vision and one from vision to language.
The query and context vectors are the outputs of the $(k\mbox{-}1)$-th layer (i.e., language features $\{h_i^{k-1}\}$ and vision features $\{v_j^{k-1}\}$):
\begin{align*}
    \hat{h}^k_i & = \mathrm{CrossAtt}_{\textsc{l}\rightarrow\textsc{r}} \left( {h}^{k-1}_i{,} \{{v}^{k-1}_1{,} \ldots , {v}^{k-1}_m\}  \right) \\
    \hat{v}^k_j & = \mathrm{CrossAtt}_{\textsc{r}\rightarrow\textsc{l}} \left({v}^{k-1}_j{,}  \{{h}^{k-1}_1{,} \ldots , {h}^{k-1}_n\} \right) 
\end{align*}
The cross-attention sub-layer is used to exchange the information and align the entities between the two modalities in order to learn joint cross-modality representations.
For further building internal connections, the self-attention sub-layers (`Self') are then applied to the output of the cross-attention sub-layer:
\begin{align*}
    \tilde{h}^k_i & = \mathrm{SelfAtt}_{\textsc{l}\rightarrow\textsc{l}} \left(\hat{h}^{k}_i, \{\hat{h}^{k}_1, \ldots , \hat{h}^{k}_n\} \right) \\
    \tilde{v}^k_j &= \mathrm{SelfAtt}_{\textsc{r}\rightarrow\textsc{r}} \left(\hat{v}^{k}_j,  \{\hat{v}^{k}_1, \ldots , \hat{v}^{k}_m\} \right) 
\end{align*}
Lastly, the $k$-th layer output $\{h^k_i\}$ and $\{v^k_j\}$ are produced by feed-forward sub-layers (`FF') on top of $\{\hat{h}^k_i\}$ and $\{\hat{v}^k_j\}$.
We also add a residual connection and layer normalization after each sub-layer, similar to the single-modality encoders.
\subsection{Output Representations}
As shown in the right-most part of Fig.~\ref{fig:model}, our LXMERT cross-modality model has three outputs for language, vision, and cross-modality, respectively.
The language and vision outputs are the feature sequences generated by the cross-modality encoder. 
For the cross-modality output, following the practice in \newcite{devlin2018bert}, we append a special token [CLS] (denoted as the top yellow block in the bottom branch of Fig.~\ref{fig:model}) before the sentence words, and the corresponding feature vector of this special token in language feature sequences is used as the cross-modality output.
\begin{table*}[]
\centering
\begin{tabular}{cccccccc}
\toprule
\multirow{2}{*}{Image Split} & 
\multirow{2}{*}{Images}
& \multicolumn{6}{c}{Sentences (or Questions) }         \\ \cmidrule(lr){3-8}
                                                                                   &                                                                             & COCO-Cap & VG-Cap & VQA  & GQA   & VG-QA & All   \\ \midrule
MS COCO - VG                                                                       & 72K                                                                         & 361K     & -      & 387K & -     & -     & 0.75M  \\
MS COCO $\cap$ VG                                                                  & 51K                                                                         & 256K     & 2.54M  & 271K & 515K  & 724K  & 4.30M \\
VG - MS COCO                                                                       & 57K                                                                         & -        & 2.85M  & -    & 556K  & 718K  & 4.13M \\ \midrule
All                                                                                & 180K                                                                        & 617K     & 5.39M  & 658K & 1.07M & 1.44M & 9.18M \\
\bottomrule
\end{tabular}
\caption{Amount of data for pre-training. Each image has multiple sentences/questions. `Cap' is caption. `VG' is Visual Genome. Since MS COCO and VG share $51$K images, we list it separately to ensure disjoint image splits. }
\vspace{-8pt}
\label{table:pretrain}
\end{table*}

\section{Pre-Training Strategies}
In order to learn a better initialization which understands connections between vision and language, we pre-train our model with different modality pre-training 
tasks on a large aggregated dataset.
\subsection{Pre-Training Tasks}\label{sec:pretrain_task}
\subsubsection{Language Task:  Masked Cross-Modality LM}
\label{sec:l_task}
On the language side, we take the masked cross-modality language model (LM) task. 
As shown in the bottom branch of Fig.~\ref{fig:pretrain}, the task setup is almost same to BERT~\cite{devlin2018bert}:
words are randomly masked with a probability of $0.15$ and the model is asked to predict these masked words.
In addition to BERT where masked words are predicted from the non-masked words in the language modality, LXMERT, with its cross-modality model architecture, could predict masked words from the vision modality as well, so as to resolve ambiguity.
For example, as shown in Fig.~\ref{fig:pretrain}, it is hard to determine the masked word `carrot' from its language context but the word choice is clear if the visual information is considered.
Hence, it helps building connections from the vision modality to the language modality, and we refer to this task as masked \emph{cross-modality} LM to emphasize this difference.
We also show that loading BERT parameters into LXMERT will do harm to the pre-training procedure in Sec.~\ref{sec:bert_lxmert} since BERT can perform relatively well in the language modality without learning these cross-modality connections.

\subsubsection{Vision Task: Masked Object Prediction}
\label{sec:vision_task}
As shown in the top branch of Fig.~\ref{fig:pretrain}, we pre-train the vision side by randomly masking objects (i.e., masking RoI features with zeros) with a probability of $0.15$ and asking the model to predict proprieties of these masked objects.
Similar to the language task (i.e., masked cross-modality LM), the model can infer the masked objects either from visible objects or from the language modality. Inferring the objects from the vision side helps learn the object relationships, and inferring from the language side helps learn the cross-modality alignments.
Therefore, we perform two sub-tasks: \textbf{RoI-Feature Regression} regresses the object RoI feature $f_j$ with L2 loss, and \textbf{Detected-Label Classification} learns the labels of masked objects with cross-entropy loss.
In the `Detected-Label Classification' sub-task, although most of our pre-training images have object-level annotations, 
the ground truth labels of the annotated objects are inconsistent in different datasets
(e.g., different number of label classes). 
For these reasons, we take detected labels output by Faster R-CNN~\cite{ren2015faster}.
Although detected labels are noisy, experimental results show that these labels contribute to pre-training in Sec.~\ref{sec:vision_analysis}.
\subsubsection{Cross-Modality Tasks}
\label{sec:x_task}
As shown in the middle-rightmost part of Fig.~\ref{fig:pretrain}, to learn a strong cross-modality representation, we pre-train the LXMERT model with 2 tasks that explicitly need both language and vision modalities.
\paragraph{Cross-Modality Matching} 
For each sentence, with a probability of $0.5$, we replace it with a mis-matched\footnote{
We take a sentence from another image as the mismatched sentence. Although the sentence and the image still have chance to match each other, this probability is very low. 
}
sentence.
Then, we train a classifier to predict whether an image and a sentence match each other.
This task is similar to `Next Sentence Prediction' in BERT~\cite{devlin2018bert}.
\paragraph{Image Question Answering (QA)}
In order to enlarge the pre-training dataset (see details in Sec.~\ref{sec:pretrain_data}), around $1/3$ sentences in the pre-training data are questions about the images.
We ask the model to predict the answer to these image-related questions 
when the image and the question are matched (i.e., not randomly replaced in the cross-modality matching task).
We show that pre-training with this image QA leads to a better cross-modality representation in Sec.~\ref{sec:QA_analysis}.

\subsection{Pre-Training Data}
\label{sec:pretrain_data}
As shown in Table.~\ref{table:pretrain}, we aggregate pre-training data from five vision-and-language datasets whose images come from MS COCO~\cite{lin2014microsoft} or Visual Genome~\cite{krishna2017visual}. 
Besides the two original captioning datasets, we also aggregate three large image question answering (image QA) datasets: VQA v2.0~\cite{antol2015vqa}, GQA balanced version~\cite{hudson2019gqa}, and VG-QA~\cite{zhu2016visual7w}.
We only collect \textbf{train and dev} splits in each dataset to avoid seeing any test data in pre-training.
We conduct minimal pre-processing on the five datasets to create aligned image-and-sentence pairs.
For each image question answering dataset, we take questions as sentences from the image-and-sentence data pairs and take answers as labels in the image QA pre-training task (described in Sec.~\ref{sec:x_task}).
This provides us with a large aligned vision-and-language dataset of $9.18$M image-and-sentence pairs on $180$K distinct images.
In terms of tokens, the pre-training data contain around $100$M words and $6.5$M image objects.

\begin{table*}[]
\centering
\begin{tabular}{lccccccccc}
\toprule
\multirow{2}{*}{Method}   & \multicolumn{4}{c}{VQA}                                                  & \multicolumn{3}{c}{GQA}                                                  & \multicolumn{2}{c}{\NLVR}    \\ 
\cmidrule(lr){2-5}
\cmidrule(lr){6-8} \cmidrule(lr){9-10}
 & Binary           & Number           & Other         & \multicolumn{1}{c}{\textbf{Accu}} & Binary           & Open                   & \multicolumn{1}{c}{\textbf{Accu}} & Cons & \textbf{Accu}          \\ \midrule
  Human     &  -  &  - & -  & -  & 91.2  &   87.4      & 89.3  & -    &96.3 \\
 Image Only       & - &   -   &-    & -      & 36.1    & 1.74      &   17.8  & 7.40    & 51.9  \\ 
 Language Only     &  66.8  &  31.8 & 27.6  & 44.3  & 61.9  &   22.7    & 41.1  &4.20    &51.1 \\
 \midrule
State-of-the-Art               & 85.8          & 53.7          & 60.7          & 70.4                      & 76.0          & 40.4          & 57.1                      &    12.0       & 53.5          \\ \midrule
LXMERT                      & \textbf{88.2} & \textbf{54.2} & \textbf{63.1} & \textbf{72.5}             & \textbf{77.8} & \textbf{45.0} & \textbf{60.3}             & \textbf{42.1} & \textbf{76.2} \\
\bottomrule
\end{tabular}
\caption{Test-set results. VQA/GQA results are reported on the `test-standard' splits and $\text{NLVR}^2$ results are reported on the unreleased test set (`Test-U'). The highest method results are in bold. Our LXMERT framework outperforms previous (comparable) state-of-the-art methods on all three datasets w.r.t. all metrics. }
\label{table:result}
\vspace{-10pt}
\end{table*}

\subsection{Pre-Training Procedure}
\label{sec:pretrain_procedure}
We pre-train our LXMERT model on the large aggregated dataset (discussed in Sec.~\ref{sec:pretrain_data}) via the pre-training tasks (Sec.~\ref{sec:pretrain_task}).
The details about the data splits are in the Appendix.
The input sentences are split by the WordPiece tokenizer~\cite{wu2016google} provided in BERT~\cite{devlin2018bert}.
The objects are detected by Faster R-CNN~\cite{ren2015faster} which is pre-trained on Visual Genome (provided by \newcite{anderson2018bottom}).
We do not fine-tune the Faster R-CNN detector and freeze it as a feature extractor.
Different from detecting variable numbers of objects in \newcite{anderson2018bottom}, we consistently keep $36$ objects for each image to maximize the pre-training compute utilization by avoiding padding.
For the model architecture, we set the numbers of layers \NL, \NX, and \NR to $9$, $5$, and $5$ respectively.\footnote{If we count a single modality layer as one half cross-modality layer, the equivalent number of cross-modality layers is $(9 + 5) / 2 + 5 = 12$, which is same as the number of layers in $\text{BERT}_\text{BASE}$.}
More layers are used in the language encoder to balance the visual features extracted from $101$-layer Faster R-CNN.
The hidden size $768$ is the same as $\text{BERT}_\text{BASE}$.
We pre-train all parameters in encoders and embedding layers from scratch
(i.e., model parameters are randomly initialized or set to zero). 
We also show results of loading pre-trained BERT parameters in Sec.~\ref{sec:bert_lxmert}.
LXMERT is pre-trained with multiple pre-training tasks and hence multiple losses are involved.
We add these losses with equal weights.
For the image QA pre-training tasks, we create a joint answer table with $9500$ answer candidates which roughly cover $90\%$ questions in all three image QA datasets.

We take Adam~\cite{kingma2014adam} as the optimizer with a linear-decayed learning-rate schedule~\cite{devlin2018bert} and a peak learning rate at $1e-4$.
We train the model for $20$ epochs (i.e., roughly $670$K\footnote{For comparison, ResNet on ImageNet classification takes $600$K steps and BERT takes $1000$K steps.} optimization steps) with a batch size of $256$.
We only pre-train with image QA task (see Sec.~\ref{sec:x_task}) for the last $10$ epochs, because this task converges faster and empirically needs a smaller learning rate. 
The whole pre-training process takes $10$ days on $4$ Titan Xp.
\paragraph{Fine-tuning}
Fine-tuning is fast and robust.
We only perform necessary modification to our model with respect to different tasks (details in Sec.~\ref{sec:implementation}).
We use a learning rate of $1e-5$ or $5e-5$, a batch size of $32$, and fine-tune the model from our pre-trained parameters for $4$ epochs.

\section{Experimental Setup and Results}
In this section, we first introduce the datasets that are used to evaluate our LXMERT framework and empirically compare our single-model results with previous best results.
\subsection{Evaluated Datasets}
\label{sec:datasets}
We use three datasets for evaluating our LXMERT framework: VQA v2.0 dataset~\cite{goyal2017making}, GQA~\cite{hudson2019gqa}, and $\text{NLVR}^2$. See details in Appendix.

\subsection{Implementation Details}
\label{sec:implementation}
On VQA and GQA, we fine-tune our model from the pre-trained snapshot without data augmentation (analysis in Sec.~\ref{sec:QA_analysis}).
When training GQA, we only take raw questions and raw images as inputs and do not use other supervisions (e.g., functional programs and scene graphs).
Since each datum in $\text{NLVR}^2$ has two natural images $\mathit{img}_0, \mathit{img}_1$ and one language statement $s$, we use LXMERT to encode the two image-statement pairs $(\mathit{img}_0, s)$ and $(\mathit{img}_1, s)$, then train a classifier based on the concatenation of the two cross-modality outputs.
More details in Appendix.
\subsection{Empirical Comparison Results}
\label{sec:results}
We compare our single-model results with previous best published results on VQA/GQA test-standard sets and $\text{NLVR}^2$ public test set.
Besides previous state-of-the-art (SotA) methods, we also show the human performance and image-only/language-only results when available.
\paragraph{VQA}
The SotA result is BAN+Counter in \newcite{kim2018bilinear}, which achieves the best accuracy among other recent works: MFH~\cite{yu2018beyond}, Pythia~\cite{jiang2018pythia}, DFAF~\cite{Gao_2019_CVPR}, and Cycle-Consistency~\cite{shah2019cycle}.\footnote{
These are state-of-the-art methods at the time of our EMNLP May 21, 2019 submission deadline. 
Since then, there have been some recently updated papers such as MCAN~\cite{yu2019deep},  MUAN~\cite{yu2019multimodal}, and MLI~\cite{gao2019multimodality}. MCAN (VQA challenge version) uses stronger mixture of detection features and achieves 72.8\% on VQA 2.0 test-standard. MUAN achieves 71.1\% (compared to our 72.5\%).
}
LXMERT improves the SotA overall \emph{accuracy} (`Accu' in Table~\ref{table:result}) by $2.1\%$ and has $2.4\%$ improvement on the `Binary'/`Other' question sub-categories.
Although LXMERT does not explicitly take a counting module as in BAN+Counter, our result on the counting-related questions (`Number') is still equal or better.\footnote{Our result on VQA v2.0 `test-dev' is 72.4\%.}
\paragraph{GQA}
The GQA~\cite{hudson2019gqa} SotA result is taken from BAN~\cite{kim2018bilinear} on the public leaderbaord.
Our $3.2\%$ \emph{accuracy} gain over the SotA GQA method is higher than VQA, possibly because GQA requires more visual reasoning. 
Thus our framework, with novel encoders and cross-modality pre-training, is suitable and achieves a $4.6\%$ improvement on open-domain questions (`Open' in Table~\ref{table:result}).\footnote{Our result on GQA `test-dev' is 60.0\%.}
\paragraph{\NLVR}
$\text{NLVR}^2$~\cite{suhr2018corpus} is a challenging visual reasoning dataset where some existing approaches~\cite{hu2017learning, perez2018film} fail, and the SotA method is `MaxEnt' in \newcite{suhr2018corpus}.
The failure of existing methods (and our model w/o pre-training in Sec.~\ref{sec:bert_lxmert}) indicates that the connection between vision and language may not be end-to-end learned in a complex vision-and-language task without large-scale pre-training.
However,  with our novel pre-training strategies in building the cross-modality connections, we significantly improve the \emph{accuracy} (`Accu' of 76.2\% on unreleased test set `Test-U', in Table~\ref{table:result}) by $22\%$.
Another evaluation metric \emph{consistency} measures the proportion of unique sentences for which all related image pairs\footnote{Each statement in $\text{NLVR}^2$ is related to multiple image pairs in order to balance the dataset answer distribution.} are correctly predicted.
Our LXMERT model improves \emph{consistency} (`Cons') to 42.1\% (i.e., by $3.5$ times).\footnote{These are the unreleased test set (`Test-U') results. On the public test set (`Test-P'), LXMERT achieves 74.5\% Accu and 39.7\% Cons.}

\begin{table}[]
\centering
\begin{tabular}{lccc}
\toprule
Method                      & VQA   & GQA   & \NLVR   \\
\midrule
LSTM + BUTD                 & 63.1  & 50.0  & 52.6 \\
BERT + BUTD                 & 62.8  & 52.1  & 51.9\\
\midrule
BERT + 1 CrossAtt           & 64.6  & 55.5  & 52.4 \\
BERT + 2 CrossAtt           & 65.8  & 56.1  & 50.9 \\
BERT + 3 CrossAtt           & 66.4  & 56.6  & 50.9 \\
BERT + 4 CrossAtt           & 66.4  & 56.0  & 50.9 \\
BERT + 5 CrossAtt           & 66.5  & 56.3  & 50.9 \\
\midrule
Train + BERT                & 65.5  & 56.2  & 50.9 \\
Train + scratch             & 65.1  & 50.0  & 50.9 \\
Pre-train + BERT            & 68.8  & 58.3  & 70.1  \\
\textbf{Pre-train + scratch}  & \textbf{69.9}  &\textbf{60.0} & \textbf{74.9}\\
\bottomrule
\end{tabular}
\vspace{-5pt}
\caption{Dev-set accuracy of using BERT. 
}
\label{table:bert}
\vspace{-5pt}
\end{table}
\section{Analysis}
In this section, we analyze our LXMERT framework by comparing it with some alternative choices or by excluding certain model components/pre-training strategies.
\subsection{BERT versus LXMERT} 
\label{sec:bert}
BERT~\cite{devlin2018bert} is a pre-trained language encoder which improves several language tasks.
As shown
in Table~\ref{table:bert}, we discuss several ways to incorporate a $\text{BERT}_\text{BASE}$ pre-trained model for vision-language tasks and empirically compare it with our LXMERT approach. 
Although our full model achieves accuracy of $74.9\%$ on $\text{NLVR}^2$, all results without LXMERT pre-training is around $22\%$ absolute lower.
\paragraph{BERT+BUTD}
\label{sec:bert_butd}
Bottom-Up and Top-Down (BUTD) attention~\cite{anderson2018bottom} method encodes questions with GRU~\cite{chung2015gated}, then attends to object RoI features $\{f_j\}$ to predict the answer.
We apply BERT to BUTD by replacing its GRU language encoder with BERT.
As shown in the first block of Table.~\ref{table:bert}, results of BERT encoder is comparable to LSTM encoder.

\paragraph{BERT+CrossAtt}
\label{sec:bert_crossatt}
Since BUTD only takes the raw RoI features $\{f_j\}$ without considering the object positions $\{p_j\}$ and object relationships, we enhance BERT+BUTD with our novel position-aware object embedding (in Sec.~\ref{sec:obj_emb}) and cross-modality layers (in Sec.~\ref{sec:crossatt}).
As shown in the second block of Table~\ref{table:bert}, the result of $1$ cross-modality layer is better than BUTD, while stacking more cross-modality layers further improves it.
However, without our cross-modality pre-training (BERT is language-only pre-trained),  results become stationary after adding $3$ cross-attention layers and have a $3.4\%$ gap to our full LXMERT framework (the last bold row in Table~\ref{table:bert}).
\paragraph{BERT+LXMERT}
\label{sec:bert_lxmert}
We also try loading BERT parameters\footnote{
Since our language encoder is same as $\text{BERT}_\text{BASE}$, except the number of layers (i.e., LXMERT has $9$ layers and BERT has $12$ layers), we load the top $9$ BERT-layer parameters into the LXMERT language encoder.
}
into LXMERT, and use it in model training (i.e., without LXMERT pre-training) or in pre-training.
We show results in the last block of Table.~\ref{table:bert}.
Compared to the `from scratch' (i.e., model parameters are randomly initialized) approach, BERT improves the fine-tuning results but it shows weaker results than our full model.
Empirically, pre-training LXMERT initialized with BERT parameters has lower (i.e., better) pre-training loss for the first $3$ pre-training epochs but was then caught up by our `from scratch' approach.
A possible reason is that BERT is already pre-trained with single-modality masked language model, and thus could do well based only on the language modality without considering the connection to the vision modality (as discussed in Sec.~\ref{sec:l_task}).

\begin{table}[]
\centering
\begin{tabular}{lccc}
\toprule
Method                      & VQA   & GQA   & $\text{NLVR}^2$ \\
\midrule
1. P20 + DA                  &68.0   &58.1   &  -              \\
2. P20 + FT                  &68.9   &58.2   &  72.4           \\
3. P10+QA10 + DA             &69.1   &59.2   &   -             \\
\textbf{4. P10+QA10 + FT}    &\textbf{69.9}  &\textbf{60.0} &  \textbf{74.9}          \\
\bottomrule
\end{tabular}
\caption{Dev-set accuracy showing the importance of the image-QA pre-training task. P10 means pre-training without the image-QA loss for $10$ epochs while QA10 means pre-training with the image-QA loss. 
DA and FT mean fine-tuning with and without Data Augmentation, resp.
}
\label{table:QA}
\vspace{-10pt}
\end{table}

\subsection{Effect of the Image QA Pre-training Task}
\label{sec:QA_analysis}
We show the importance of image QA pre-training task (introduced in Sec.~\ref{sec:x_task}) by excluding it or comparing it with its alternative: data augmentation.
\paragraph{Pre-training w/ or w/o Image QA}
To fairly compare with our original pre-training procedure (10 epochs w/o QA + 10 epochs w/ QA, details in Sec.~\ref{sec:pretrain_procedure}) , we pre-train LXMERT model without image QA task for $20$ epochs.
As shown in Table~\ref{table:QA} rows 2 and 4, pre-training with QA loss improves the result on all three datasets. 
The $2.1\%$ improvement on $\text{NLVR}^2$ shows the stronger representations learned with image-QA pre-training, since all data (images and statements) in $\text{NLVR}^2$ are not used in pre-training.

\begin{table}[]
\centering
\begin{tabular}{lccc}
\toprule
Method                                & VQA   & GQA  & $\text{NLVR}^2$ \\
\midrule
1. No Vision Tasks                     & 66.3  & 57.1 & 50.9 \\
2. Feat                                & 69.2  & 59.5 & 72.9 \\
3. Label                               & 69.5  & 59.3 & 73.5 \\
\textbf{4. Feat + Label}               & \textbf{69.9}  & \textbf{60.0} & \textbf{74.9} \\
\bottomrule
\end{tabular}
\vspace{-5pt}
\caption{Dev-set accuracy of different vision pre-training tasks. `Feat' is RoI-feature regression; `Label' is detected-label classification. }
\vspace{-10pt}
\label{table:vision}
\end{table}

\paragraph{Pre-training versus Data Augmentation}
Data augmentation (DA) is a technique which is used in several VQA implementations~\cite{anderson2018bottom, kim2018bilinear, jiang2018pythia}.
It increases the amount of training data by adding questions from other image QA datasets.
Our LXMERT framework instead uses multiple QA datasets in pre-training and is fine-tuned only on one specific dataset.
Since the overall amounts of data used in pre-training and DA are similar, we thus can fairly compare these two strategies, and results show that our QA pre-training approach outperforms DA.
We first exclude the QA task in our pre-training and show the results of DA fine-tuning.
As shown in Table.~\ref{table:QA} row 1, DA fine-tuning decreases the results compared to non-DA fine-tuning in row 2.
Next, we use DA after QA-pre-training (row 3) and DA also drops the results.

\subsection{Effect of Vision Pre-training tasks}
\label{sec:vision_analysis}
We analyze the effect of different vision pre-training tasks in Table~\ref{table:vision}.
Without any vision tasks in pre-training (i.e., only using the language and cross-modality pre-training tasks), the results (row $1$ of Table~\ref{table:vision}) are similar to BERT+3 CrossAtt in Table \ref{table:bert}.
The two visual pre-training tasks (i.e., RoI-feature regression and detected-label classification) could get reasonable results (row $2$ and row $3$) on their own, and jointly pre-training with these two tasks achieves the highest results (row $4$).

\subsection{Visualizing LXMERT Behavior}
In the appendix, we show the behavior of LXMERT by visualizing its attention graphs in the {language encoder}, {object-relationship encoder}, and {cross-modality encoder}, respectively.
\section{Related Work}
\paragraph{Model Architecture} Our model is closely related to three ideas: bi-directional attention, Transformer, and BUTD.
\newcite{lu2016hierarchical} applies bi-directional attention to the vision-and-language tasks while its concurrent work BiDAF~\cite{seo2016bidirectional} adds modeling layers in solving reading comprehension.
Transformer~\cite{vaswani2017attention} is first used in machine translation, we utilize it as our single-modality encoders and design our cross-modality encoder based on it.
BUTD~\cite{anderson2018bottom} embeds images with the object RoI features, we extend it with object positional embeddings and object relationship encoders.

\paragraph{Pre-training} After ELMo~\cite{peters2018deep}, GPT~\cite{radford2018improving}, and BERT~\cite{devlin2018bert} show improvements in language understanding tasks with large-scale pre-trained language model, progress has been made towards the cross-modality pre-training.
XLM~\cite{lample2019cross} learns the joint cross-lingual representations by leveraging the monolingual data and parallel data.
VideoBert~\cite{sun2019videobert} takes masked LM on the concatenation of language words and visual tokens, where the visual tokens are converted from video frames by vector quantization.
However, these methods are still based on a single transformer encoder and BERT-stype token-based pre-training, thus we develop a new model architecture and novel pre-training tasks to satisfy the need of cross-modality tasks.

\paragraph{Recent works since our EMNLP submission}
This version of our paper (and all current results) was submitted to EMNLP\footnote{EMNLP deadline was on May 21, 2019, and the standard ACL/EMNLP arxiv ban rule was in place till the notification date of August 12, 2019.} and was used to participate in the VQA and GQA challenges in May 2019.
Since our EMNLP submission, a few other useful preprints have recently been released (in August) on similar cross-modality pre-training directions: ViLBERT~\cite{lu2019vilbert} and VisualBERT~\cite{li2019visualbert}.
Our LXMERT methods differs from them in multiple ways: we use a more detailed, multi-component design for the cross-modality model (i.e., with an object-relationship encoder and cross-modality layers) and we employ additional, useful pre-training tasks (i.e., RoI-feature regression and image question answering).
These differences result in the current best performance (on overlapping reported tasks): 
a margin of 1.5\% accuracy on VQA 2.0 and a margin of 9\% accuracy on $\text{NLVR}^2$ (and 15\% in consistency). LXMERT is also the only method which ranks in the top-3 on both the VQA and GQA challenges among more than 90 teams.
We provide a detailed analysis to show how these additional pre-training tasks contribute to the fine-tuning performance in Sec.~\ref{sec:QA_analysis} and Sec.~\ref{sec:vision_analysis}.

\section{Conclusion}
We presented a cross-modality framework, LXMERT, for learning the connections between vision and language.
We build the model based on Transfermer encoders and our novel cross-modality encoder.
This model is then pre-trained with diverse pre-training tasks on a large-scale dataset of image-and-sentence pairs.
Empirically, we show state-of-the-art results on two image QA datasets (i.e., VQA and GQA) and show the model generalizability with a $22\%$ improvement on the challenging visual reasoning dataset of $\text{NLVR}^2$.
We also show the effectiveness of several model components and training methods via detailed analysis and ablation studies.

\section*{Acknowledgments}
We thank the reviewers for their helpful comments. This work was supported by ARO-YIP Award \#W911NF-18-1-0336, and awards from Google, Facebook, Salesforce, and Adobe. The views, opinions, and/or findings contained in this article are those of the authors and should not be interpreted as representing the official views or policies, either expressed or implied, of the funding agency. We also thank Alane Suhr for evaluation on NLVR$^2$. 

\bibliography{emnlp-ijcnlp-2019}
\bibliographystyle{acl_natbib}

\appendix

\section*{Appendix}

\section{Evaluated Datasets Description}
We use three datasets for evaluating our LXMERT framework. 
\paragraph{VQA}
The goal of visual question answering (VQA)~\cite{antol2015vqa} is to answer a natural language question related to an image.
We take VQA v2.0 dataset~\cite{goyal2017making} which reduces the answer bias compared to VQA v1.0.
The dataset contains an average of $5.4$ questions per image and the total amount of questions is $1.1$M.
\paragraph{GQA}
The task of GQA~\cite{hudson2019gqa} is same as VQA (i.e., answer single-image related questions), but GQA requires more reasoning skills (e.g., spatial understanding and multi-step inference).
$22$M questions in the dataset are generated from ground truth image scene graph to explicitly control the question quality.
\paragraph{\NLVR}
Since the previous two datasets are used in pre-training for increasing the amount of pre-training data to a certain scale, we evaluate our LXMERT framework on another challenging visual reasoning dataset $\text{NLVR}^2$ where all the sentences and images are not covered in pre-training.
Each datum in $\text{NLVR}^2$ contains two related natural images and one natural language statement. 
The task is to predict whether the statement correctly describes these two images or not.
$\text{NLVR}^2$ has $86$K, $7$K, $7$K data in training, development, and test sets, respectively.

\section{Details of $\text{NLVR}^2$ Fine-tuning}
Each datum in $\text{NLVR}^2$ consists of a two-image pair ($img_0$, $img_1$), one statement $s$, and a ground truth label $y^*$ indicating whether the statement correctly describe the two images.
The task is to predict the label $y$ given the images and the statement.

To use our LXMERT model on $\text{NLVR}^2$, we concatenate the cross-modality representations of the two images and then build the classifier with GeLU activation\cite{hendrycks2016bridging}.
Suppose that $\mathrm{LXMERT}(\mathit{img}, \mathit{sent})$ is the single-vector cross-modality representation, the predicted probability is:
\begin{align*}
x_0 &= \mathrm{LXMERT}(\mathit{img}_0, s) \\
x_1 &= \mathrm{LXMERT}(\mathit{img}_1, s) \\
z^0 &= W_0 [x_0; x_1] + b_0 \\
z^1 &= \mathrm{LayerNorm}\left(\mathrm{GeLU}(z^0)\right) \\
\mathit{prob} &= \sigma(W_1 z^1 + b_1)
\end{align*}
where $\sigma$ is sigmoid function.
The model is optimized by maximizing the log-likelihood, which is equivalent to minimize the binary cross entropy loss:
\begin{align*}
    \mathcal{L} = \mbox{-} y^* \log \mathit{prob} - (1 - y^*) \log (1- \mathit{prob})
\end{align*}

\section{Training, Validation, and Testing Splits}
We carefully split each dataset to ensure that all testing images are not involved in any pre-training or fine-tuning steps.
Our data splits for each dataset and reproducible code are available at \href{https://github.com/airsplay/lxmert}{https://github.com/airsplay/lxmert}.
\paragraph{LXMERT Pre-Traininig}
Since MS COCO has a relative large validation set, 
we sample a set of $5$k images from the MS COCO validation set as the mini-validation (minival) set.
The rest of the images in training and validation sets (i.e., COCO training images, COCO validation images besides minival, and all the other images in Visual Genome) are used in pre-training.
Although the captions and questions of the MS COCO test sets are available, we exclude all of them to make sure that testing images are not seen in pre-training.
\paragraph{Fine-tuning}
For training and validating VQA v2.0, we take the same split convention as in our LXMERT pre-training.
The data related to images in LXMERT mini-validation set is used to validate model performance and the rest of the data in train+val are used in fine-tuning.
We test our model on the VQA v2.0 `test-dev' and `test-standard' splits.
For GQA fine-tuning, we follow the suggestions in official GQA guidelines\footnote{
\href{https://cs.stanford.edu/people/dorarad/gqa/evaluate.html}{https://cs.stanford.edu/people/dorarad/gqa/evaluate.html}
} to take \emph{testdev} as our validation set and fine-tune our model on the joint train + validation sets.
We test our GQA model on GQA `test-standard' split.
The images in $\text{NLVR}^2$ are not from either MS COCO or Visual Genome, we thus keep using the original split: fine-tune on train split, validate the model choice on val split, and test on the public (`Test-P') and unreleased (`Test-U') test splits.

\section{Training Details of `BERT versus LXMERT'}
When training with BERT only, we train each experiments for $20$ epochs with a batch size $64$/$128$ since it was not pre-trained on these cross-modality datasets.
The learning rate is set to $1e-4$ instead of $5e-5$.

\begin{figure}[t]
\centering
\includegraphics[width=0.47\textwidth]{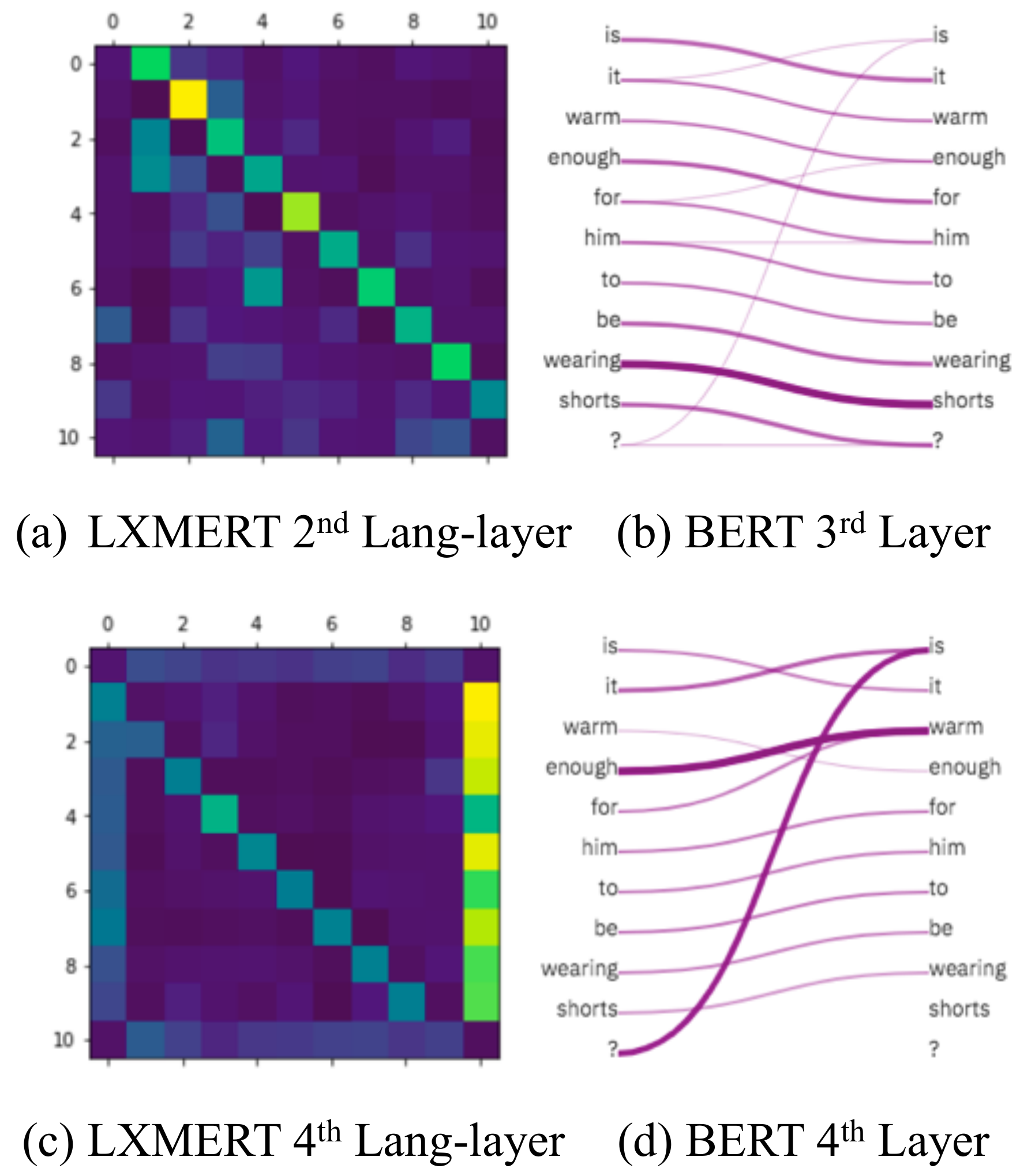}
\caption{
Attention graphs reveal similar behavior in the LXMERT language encoder (a, c) and in the original BERT encoder (b, d). Fig. a \& b show the attention pointing to next words while Fig. c \& d show the attention pointing to previous words.
}
\vspace{0pt}
\label{fig:lang_att}
\end{figure}

\begin{figure}[t]
\centering
\includegraphics[width=0.47\textwidth]{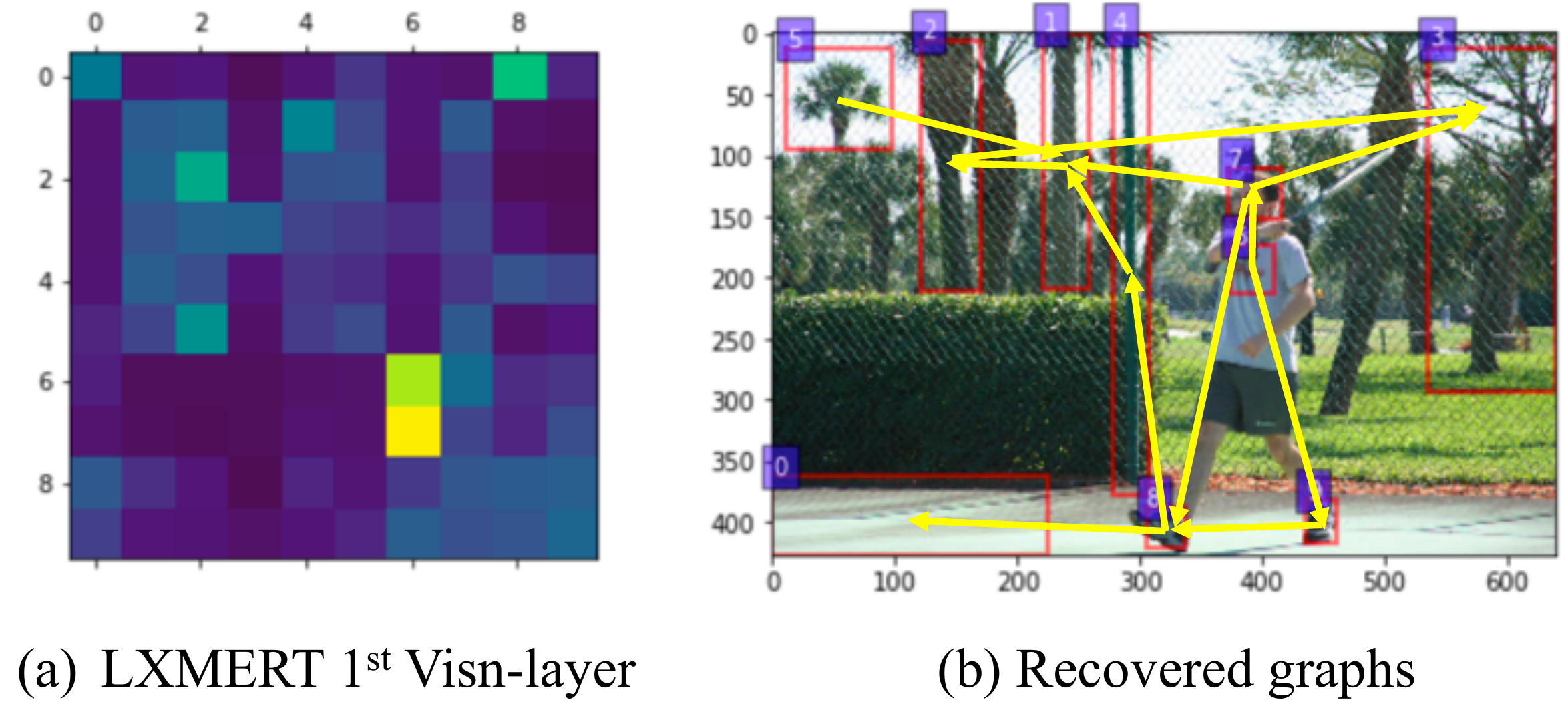}
\caption{
The attention graph (a) and its recovered scene graph (b) in the first layer of LXMERT's object-relationship encoder.
}
\vspace{0pt}
\label{fig:visn_att}
\end{figure}

\section{Visualizing LXMERT Behavior}
In this section, we show the behavior of LXMERT by visualizing its attention graphs in the {language encoder}, {object-relationship encoder}, and {cross-modality encoder}, respectively.
\subsection{Language Encoder}
In Fig.~\ref{fig:lang_att}, we reveal that the LXMERT language encoder has similar behaviour as the original BERT encoder, by using the same sentence ``Is it warm enough for him to be wearing shorts?'' as the input to both models.
LXMERT's attention graphs (in Fig.~\ref{fig:lang_att}(a, c)) are extracted from the pre-trained LXMERT without fine-tuning on a specific task.
BERT's attention graphs (in Fig.~\ref{fig:lang_att}(b, d)) come from \newcite{hoover2019exbert}.\footnote{exBERT demo \cite{hoover2019exbert} is available at \url{http://exbert.net/}}
We find that both the second LXMERT layer (Fig.~\ref{fig:lang_att}(a)) and third BERT layer (Fig.~\ref{fig:lang_att}(b)) point to the next words while both the fourth LXMERT layer (Fig.~\ref{fig:lang_att}(c)) and fourth BERT layer (Fig.~\ref{fig:lang_att}(d)) point to the previous words, thus showing the similar behaviour of the two encoders.

\begin{figure}[t]
\centering
\includegraphics[width=0.47\textwidth]{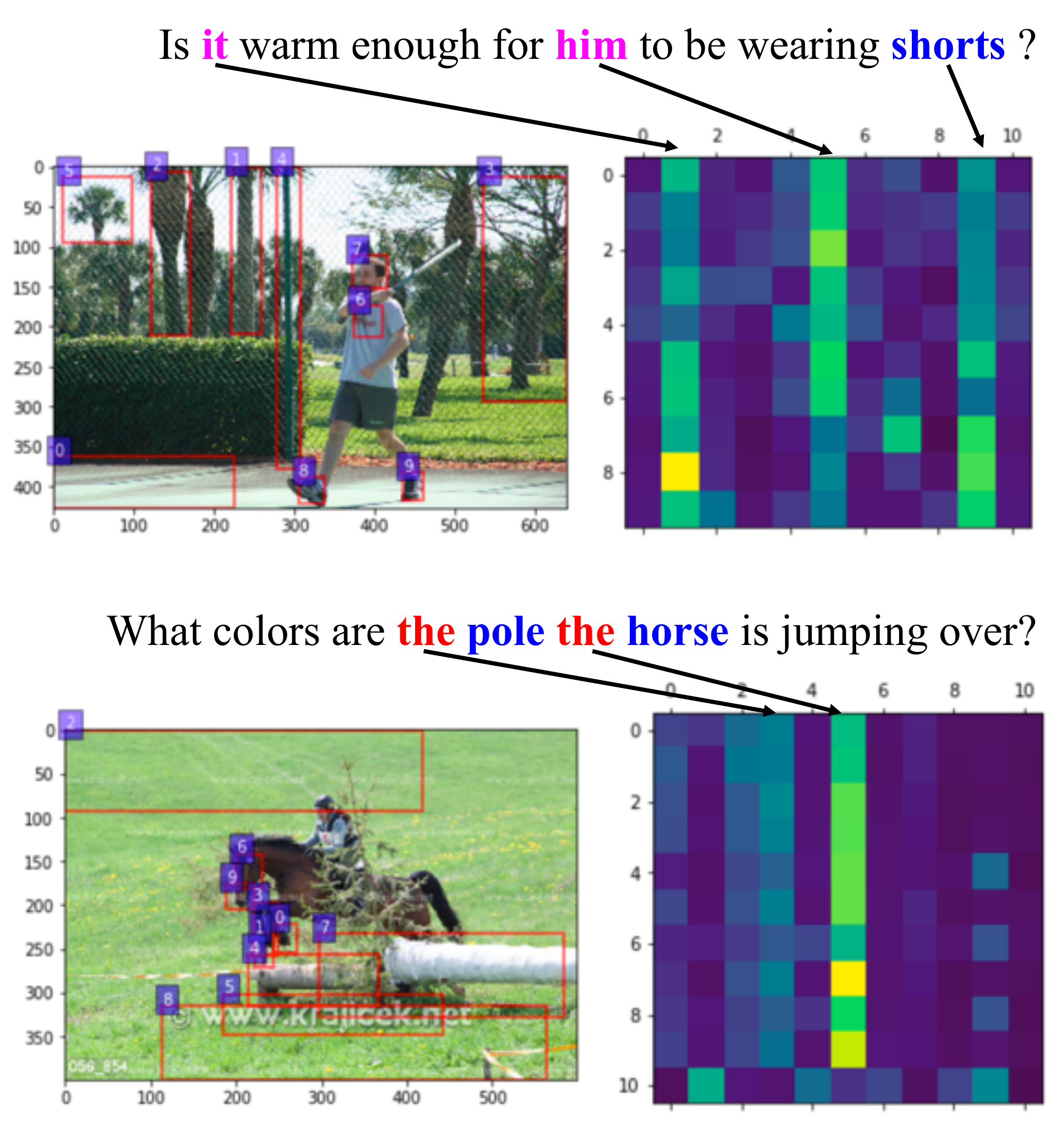}
\caption{
Attention graphs in LXMERT's cross-modality encoder showing that the attention focus on pronouns (marked in pink), nouns (marked in blue), and articles (marked in red). 
}
\vspace{-5pt}
\label{fig:cross_att}
\end{figure}

\subsection{Object-Relationship Encoder}
In Fig.~\ref{fig:visn_att}, we visualize the attention graph of the first layer in LXMERT's object-relationship encoder. 
We only highlight the objects with the highest attention scores while the other objects are mostly not attended to.
We manually build the connections between objects (marked as yellow lines in Fig.~\ref{fig:visn_att}(b)) according to the attention graph.
These connections faithfully draw a scene graph of the figure, which indicates that the object-relationship encoder might be learning a reasonably good network of the relationships between objects.

\subsection{Cross-Modality Encoder}
In Fig.~\ref{fig:cross_att}, we visualize the attention in LXMERT's cross-modality encoder to reveal the connections between objects and words. 
We find that the attention focuses on nouns and pronouns as shown in the top figure of Fig.~\ref{fig:cross_att} because they are the most informative words in current vision-and-language tasks.
However, for non-plural nouns (as shown in the bottom example in Fig.~\ref{fig:cross_att}), the attention will focus on the articles.
Although we do not specifically design for this behavior, we think that articles are possibly serving as special tokens (e.g., [CLS], [SEP] in BERT), thus providing unified target entries for the attention layers. Next, we are also looking at how to utilize pre-training tasks which directly capture pairwise noun-noun and noun-verb relationships between the images and text sentences.

\end{document}